\documentclass[paper]{ieice}
\usepackage[pdftex]{graphicx,xcolor}
\usepackage[fleqn]{amsmath}
\usepackage{newtxtext}
\usepackage[varg]{newtxmath}
\usepackage{booktabs}
\usepackage{float}
\usepackage{xspace}
\usepackage{cleveref}
\usepackage{latexml}
\crefname{figure}{Fig.}{Figs.}
\Crefname{figure}{Fig.}{Figs.}
\crefname{table}{Table}{Tables}
\Crefname{table}{Table}{Tables}
\crefname{section}{Section}{Sections}
\Crefname{section}{Section}{Sections}
\crefname{equation}{Eq.}{Eqs.}
\Crefname{equation}{Eq.}{Eqs.}
\crefname{footnote}{Footnote}{Footnotes}
\Crefname{footnote}{Footnote}{Footnotes}

\def\heading#1{\noindent\textbf{#1}}

\newcommand{\etal}{\emph{et al.}\xspace}

\title{Projector Time Map Calibration for Event-based Structured Light}

\iflatexml
  \author{Buu-Hoang-Lam Le \quad Kota Mizutani \quad Shohei Nobuhara \\
    {Department of Information Science, Kyoto Institute of Technology}}
  \newenvironment{summary}{\begin{abstract}}{\end{abstract}}
\else
  \vol{110}
  \no{1}
  \field{D}
  \authorlist{%
   \authorentry{Buu-Hoang-Lam Le}{n}{a}\MembershipNumber{}
   \authorentry{Kota Mizutani}{n}{a}\MembershipNumber{}
   \authorentry{Shohei Nobuhara}{m}{a}\MembershipNumber{0728080}
  }
  \affiliate[a]{The authors are with the Department of Information Science,
    Kyoto Institute of Technology.}
  \received{2026}{1}{1}
  \revised{2026}{1}{1}
\fi

\begin{document}
\maketitle
\begin{summary}
Existing event-based structured light obtains a dense projector-camera
  correspondence from the timestamp of an event.
  It is commonly computed by assuming that the projector scans a parallel
  raster and that the sensor timestamps each pixel independently, and we show
  that neither assumption holds.
  We obtain the correspondence from a temporal code that identifies the
  projector pixel directly and therefore needs no prior calibration, we
  calibrate the projector time map from the same recording, and we recover the
  geometry of the system and the offsets between its clocks by a
  spatio-temporal self-calibration in a preamble of a few seconds that needs no
  reference object and no second camera.
\end{summary}

\section{Introduction}\label{sec:intro}

Dense 3D shape measurement at both high spatial resolution and high frame rate
is an enabling technology for the quantitative analysis, modeling, and synthesis
of shape changes that appear only briefly, such as the micro-expressions of a
human face.
Such a measurement has to produce a depth value at every pixel, to be accurate at
the scale of the change of interest, and to be completed within one frame.
Existing systems satisfy these three requirements only at a high cost.
The Mugsy v2 stage of Meta Reality Labs, for example, acquires facial geometry
with 160 cameras at 4K and 90\,Hz~\cite{wuu2023multiface}.
Depth sensors that are in wide use have other limits.
LiDAR provides only a sparse point cloud, time-of-flight cameras do not reach
millimeter accuracy~\cite{sarbolandi2015kinect}, passive stereo and active stereo
return a depth that is smoothed by the size of the matching
window~\cite{zhang2012kinect,keselman2017realsense}, and a projector-camera
system must spend either time or image area on its code: a temporal code needs a
sequence of patterns~\cite{inokuchi1984range}, and a single-shot spatial code is
measured only where the codeword lies, so a dense result rests on a
prior~\cite{furukawa2022single}.

\begin{figure}[t]
  \centering
  \begin{minipage}[c]{0.315\linewidth}\centering
    \includegraphics[height=4.4cm]{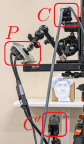}
  \end{minipage}\hfill
  \begin{minipage}[c]{0.665\linewidth}\centering
    \includegraphics[width=\linewidth]{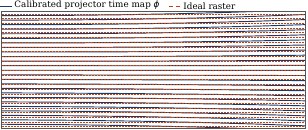}\\[2pt]
    \includegraphics[width=\linewidth]{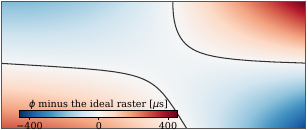}
  \end{minipage}
  \caption{Left: the capture setup, with the laser-scanning projector $P$, the
    measuring camera $C$ and the verification camera $C'$.
    Top right: iso-contours of the calibrated time map $\phi$ over the projector
    image, one every 0.5\,ms, and the rows at which an ideal raster would place
    the same instants. The contours are tilted, the tilt changes with the row,
    and the two families separate by as much as 26 projector rows.
    Bottom right: $\phi$ minus the ideal raster. It reaches 457\,$\mu$s, whereas
    $\phi$ itself fits the measurements to 5.07\,$\mu$s rms.
  }
  \label{fig:overview}
\end{figure}

Event cameras combined with a laser-scanning projector make such a method
possible.
The scan illuminates each projector pixel at a distinct time, and an event camera
reports each of those illuminations as a positive event with a temporal
resolution of microseconds.
The correspondences they form are dense, because the scan covers every projector
pixel, and they are obtained from a single scan of the projector.
The code is therefore carried by time and still fits inside one frame, so it
costs neither a sequence of patterns nor image area.
Several 3D measurement systems use this
principle~\cite{brandli2014adaptive,matsuda2015mc3d,Muglikar213DV,morgenstern2023x}.

These systems are commonly built on an idealized model of the hardware on both
sides of the correspondence: the projector is assumed to scan a parallel raster,
so that a time instant corresponds to a projector coordinate, and the camera to
assign an independent timestamp to each pixel, so that the timestamp of an event
is when the pixel was illuminated.
Neither assumption holds, and neither failure is specific to particular devices:
the first follows from how a resonant scanner is built, the second from the
row-wise readout that a high-resolution event sensor requires.
We measure both on our system.

The scan of the laser-scanning projector is not the raster that is assumed.
Its lines are not parallel but form a fan, and consecutive lines are scanned in
opposite directions.
Both follow from how these projectors are
built~\cite{hornstein2010enhanced,hornstein2012maradin,chen2026dualaxis}.
The fast axis is a mirror at resonance, the slow axis keeps moving while a line
is written, and straight parallel rows are reported only for a design that
steps the slow axis under position feedback.
We measure the fan on three units of the projector, which agree on its shape
and differ in its parameters (\cref{sec:eval-timemap}).
The instant at which each pixel is illuminated is described by the projector
time map~\cite{Muglikar213DV}, and the real scan departs from the idealized one
by much more than the precision the same system reaches once that map is
calibrated (\cref{fig:overview}).

The camera deviates as well.
Many events carry the same timestamp, so a timestamp names a set of projector
pixels and not one of them.
A method that reads a single projector pixel out of a timestamp therefore needs
something the sensor does not measure, the instant at which that one pixel was
illuminated.
The number of events sharing a timestamp also varies far more than the scan can
account for, so the sensor produces the variation.
\Cref{sec:eval-timemap} reports both measurements.

This paper models both deviations and designs the measurement so that they do
not degrade the result: we calibrate the projector time map explicitly, as a
smooth map over the projector image, and synchronize the event cameras with the
projector using temporally coded correspondences.
The projector time map itself has been proposed in existing
methods~\cite{Muglikar213DV,morgenstern2023x}, where it is either assumed to be
an idealized raster or measured from a plane set up for the purpose.
What we add is a calibration that runs on the recording being measured.
Our contributions are as follows.

\begin{itemize}
  \item A calibration of the projector time map, from the recording being
    measured. We also measure how far the real scan deviates from the idealized
    raster, and how much that degrades the reconstructed depth.
  \item Projector-camera correspondences that are obtained without any prior
    calibration. The temporal code identifies the projector pixel directly, so
    the correspondences require no geometric calibration, no time
    synchronization, no reference object, and no second camera.
  \item A spatio-temporal self-calibration, which recovers the geometry of the
    system and the offsets between its clocks at once. It runs as a preamble at
    the start of each capture session and needs a few seconds.
\end{itemize}

\section{Related Work}
\label{sec:related}

\heading{Event-based 3D sensing.}
An event camera records the pixel at which a brightness change occurred and the
time at which it was detected, so a spatially dense measurement requires that a
change be induced at every pixel.
There are three families of methods.
The first moves the camera continuously over a fixed
scene~\cite{hidalgo2020learning,huang2025inceventgs}, the second combines the
event stream with another sensor~\cite{zou2025highspeeddynamic3dimaging}, and
the third uses an active light source that generates events over the whole
object~\cite{brandli2014adaptive,matsuda2015mc3d,Muglikar213DV,morgenstern2023x,yu2024eventps,kitazawa2025ps,liang2025eventups,li2025active}.
Within it, photometric methods vary the direction of the illumination rather
than its position~\cite{yu2024eventps,kitazawa2025ps,liang2025eventups}, and Li
\etal~\cite{li2025active} perform active stereo between two event cameras under
an infrared pattern.
This work belongs to that family and uses a laser-scanning projector.

\heading{Structured light with a laser-scanning projector.}
Closest to this work are systems in which the projector sweeps a spot or a line
across the scene while an event camera records the events.
Brandli \etal~\cite{brandli2014adaptive} extract a pulsed laser line from the
events for terrain reconstruction.
MC3D~\cite{matsuda2015mc3d} scans a single spot and pairs each event with the
projector position that produced it, so the correspondence comes from time and
not from appearance.
ESL~\cite{Muglikar213DV} obtains the correspondence from the event time together
with the projector's scan timing, and X-maps~\cite{morgenstern2023x} turns it
into a lookup from event time to projector coordinate, which makes depth
available in real time.
All of these methods need the time at which the projector illuminates each of
its pixels.
ESL takes it from an idealized raster in which the time depends on the row
alone, and the accuracy of that model limits the accuracy of the
reconstruction.
X-maps measures the map instead, by projecting white onto a plane set parallel
to the two image planes and carrying the timestamps the camera records into the
projector image by the projective transformation fitted to the four corners of
the projected frame.
We measure it as well, but from coded correspondences in the recording being
measured, so that neither a plane nor a separate calibration recording is
required.

\heading{Calibration of projector-camera systems.}
The calibration of a projector-camera system is well established for frame-based
cameras.
The usual approach treats the projector as a \textit{reverse} camera.
The standard two-view calibration then applies as soon as correspondences are
available~\cite{depiero19963,moreno2012calibration}.
The calibration object can be dispensed with as well, because the geometry
follows from the projected pattern
itself~\cite{furukawa2009laser,yamazaki2011simultaneous}.
We follow this approach in \cref{sec:selfcalib}.
These methods, however, calibrate the geometry alone.
An event camera measures the time of a brightness change, so the instant at
which each projector pixel is illuminated enters the measurement directly and
has to be calibrated with the geometry, which is what this paper adds.

\section{Proposed Method}

\subsection{Projector-to-Event-Camera Correspondence by Temporal Coding}
\label{sec:tempcode}

We obtain the correspondence between projector pixels and event camera pixels by
a temporal code.
The projector displays a set of small features, each switched on and off over
successive frames so that the sequence encodes its position in the projector
image.
A feature is a square patch of projector pixels switched as one, and it stands
for the projector pixel at its center.
The camera decodes the blinking and recovers the projector pixel that produced
each feature together with the time at which it was lit, which
\cref{sec:timemap} and \cref{sec:capture} use.
Nothing below depends on how the features are arranged, and the arrangement used
here is given in \cref{sec:setup}.

\heading{Code.}
The codeword of a feature is an identifier of its position followed by an
error-detecting code computed from that identifier, transmitted one bit per
projector frame and repeated several times.
The code does not have to be self-clocking: the projector re-illuminates every
pixel once per frame, so a feature in its on state produces events at the frame
rate whatever the code is, and the frame clock follows from those events.
We therefore transmit the codeword without line coding.
A feature is accepted only when the error-detecting code is satisfied and when
the same identifier is decoded from more than one repetition.

\begin{figure}[tb]
  \centering
  \makebox[\linewidth]{%
    \includegraphics[height=2.2cm]{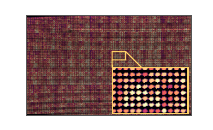}\hfill
    \includegraphics[height=2.2cm]{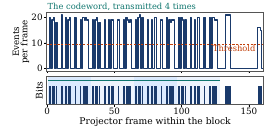}}
  \caption{The temporal code as it is received, from $R_1$ of
    \cref{tab:recordings}.
    Left: the events of one block accumulated into an image of the measuring
    camera, with a magnified region in which the individual dots and the grid
    they lie on are resolved. A dot is brighter when its codeword carries more
    one bits, and the faint ring around it is the blurred edge of its image,
    where the step in brightness reaches the threshold in some frames only.
    Right: one dot of that block. The number of events it produces in each
    projector frame is either about twenty or none, so a single threshold
    recovers one bit per frame, and all four repetitions of the codeword
    reproduce the transmitted 32 bits exactly. The frames after them carry the
    white and dark phases between blocks.
  }
  \label{fig:pattern}
\end{figure}

The features have to be separated in the camera image to be detected
individually, so one transmission, which we call a \emph{block}, provides
correspondences at a subset of the projector pixels.
A denser set is obtained by transmitting further blocks with the features
displaced, and how many are needed depends on what the correspondences are used
for, the two cases here being \cref{sec:timemap} and \cref{sec:capture}.
A block is preceded and followed by short all-white and dark phases, which we use
to segment the recording (\cref{fig:pattern}).

\heading{Decoding.}
Decoding has two parts.
The first establishes the timing of the projector frames in the clock of the
camera, because the code is sampled once per projector frame; the second is
local to each feature, whose events are then sampled frame by frame and the
resulting bit sequence decoded.
The frame timing is needed only within one frame: an error of a whole number of
frames shifts the sampled sequence without changing it, and the sliding window
below absorbs such a shift.

The frame timing is obtained in three steps.
We first segment the recording into blocks.
The pattern leaves an all-dark phase between one block and the next, so a block
appears in the event stream as a long lit interval between two dark ones, and
counting the events in short bins separates them.
We then estimate the refresh period $T$ of the projector, using the fact that
every pixel is re-illuminated once per frame so that the event times are
periodic,
\begin{equation}
  \hat{T} = \arg\max_{T}
  \left| \sum_{k=1}^{K} \exp\!\left( \frac{2 \pi \mathrm{i}\, t_k}{T} \right)
  \right| ,
  \label{eq:period}
\end{equation}
where $t_k$ is the timestamp of the $k$-th of the $K$ events used.
Each event contributes a unit vector at its phase within a period of length $T$,
so the sum is largest when successive periods reinforce one another, which they
do when $T$ is the period at which the event rate repeats.
The maximization is carried out over the whole recording and then within each
block, over a range of less than one third around the nominal period, because
the criterion is also maximal at the sub-harmonics $T/2$, $T/3$ and below.
We finally try a few candidate origins spread over one frame and keep the one
that decodes the most features.

The second part is applied to each block in turn (\cref{fig:timeline}b, c).
Features are detected by accumulating the events of the block into an image,
thresholding it and taking connected components, and a feature's position in the
camera is the centroid of its component.
For every feature and frame we count the events assigned to it and call the
feature on when the count exceeds a threshold derived from that same feature,
which gives one sampled bit per feature and frame.

The transmitted codeword may start at any of the sampled bits, so we slide a
window of the length of the codeword over the sequence and accept a window when
the error-detecting code is satisfied and the decoded identifier lies inside the
range of valid positions.
We then require at least two agreeing windows, because the number of windows
tested in one block is large enough that a single one would admit a number of
wrong identifiers that is not negligible, whereas a correct feature produces one
accepted window per repetition.
\Cref{sec:setup} reports the code used in this work.

The output of the decoding is, for every accepted feature, the projector pixel
that emitted it, the position of the feature in the camera, and the phase of the
feature inside the refresh period.

\subsection{Projector Time Map Calibration}
\label{sec:timemap}

The \emph{projector time map} is the function
\begin{equation}
  \phi : (p_x, p_y) \mapsto t \in [0, T)
  \label{eq:timemap}
\end{equation}
that returns the instant, within one refresh period $T$, at which the projector
illuminates its pixel $(p_x, p_y)$.
Every method that obtains a projector coordinate from an event timestamp uses
such a function~\cite{Muglikar213DV,morgenstern2023x}.
ESL takes it to be the timing of an ideal raster scan, the affine form
$\phi(p_x, p_y) = a + b\,p_x + c\,p_y$, which describes a parallel raster
always scanned in the same direction and at a constant rate along the line.
X-maps measures $\phi$ on a plane, as \cref{sec:related} describes, and is the
closest work to ours.
What we propose is not the time map itself but a calibration that needs no
plane and no recording of its own.
We measure $\phi$ from the same recording that transmits the code, and model it
as a smooth function over the projector image.

\heading{Properties of the real scan.}
Three properties of the actual scan, as verified in \cref{sec:eval-timemap},
determine how $\phi$ has to be modeled.
First, the scan lines are not parallel.
Their tilt changes monotonically from one edge of the frame to the other, and
they form a pencil of lines that converges at a finite vertex outside the
projected image.
A parallel raster is the special case in which this vertex is at infinity.
Second, consecutive lines are scanned in opposite directions, which is how this
class of projector is driven~\cite{hornstein2010enhanced,ultimems}.
The value of $\phi$ along one line therefore increases with $p_x$ on even lines
and decreases with $p_x$ on odd lines.
Third, one line is scanned in less time than the line period, because only the
part of the sweep where the rate is usable carries
pixels~\cite{chen2026dualaxis}.

\heading{Model.}
The fan is a smooth deformation of the ramp, but the alternation is not: no
continuous function of $(p_x, p_y)$ changes sign between two adjacent lines.
We therefore model the smooth part alone and omit the alternation.
The measurement of \cref{sec:capture} never uses the direction: a timestamp
identifies a segment of a scan line, and the point within it comes from the
epipolar constraint.
The error the omission leaves also points the harmless way: fitting the average
of the two directions is wrong by at most half the time a line takes to sweep,
6\,$\mu$s here, against the 17.5\,$\mu$s by which $\phi$ changes from one
line to the next, so the iso-timestamp contour is displaced by about a third of a
scan line rather than along it.
The model is a bivariate cubic polynomial with ten coefficients in $x$ and $y$,
the projector coordinates $p_x$ and $p_y$ each mapped linearly onto $[-1,1]$
over the projector image,
\begin{equation}
  \phi(p_x,p_y) = \sum_{i+j \le 3} c_{ij}\, x^{i} y^{j}.
  \label{eq:cubic}
\end{equation}

\heading{Estimation.}
The coefficients are obtained from a recording of the temporal code of
\cref{sec:tempcode}.
The decoding yields, for every dot, the projector pixel that emitted it, and
folding the event times of that dot modulo the refresh period recovers the phase
at which the dot was illuminated.
Each dot therefore contributes one pair $\bigl((p_x,p_y),\, \phi\bigr)$, which is
one linear equation in the ten coefficients of \cref{eq:cubic}, so the
coefficients follow from linear least squares.
The time map describes the projector alone, so this recording is made once; we
transmit as many blocks as cover the projector image densely, and one such
recording provides several hundred thousand pairs.

One nuisance parameter remains.
The phase of a block can only be referred to the period measured for that block,
and those periods differ slightly, so an unknown constant per block remains.
We estimate these constants together with the coefficients.
The constant term of $\phi$ is not determined by these observations either,
because adding a constant to it and subtracting the same constant from every
block leaves them unchanged.
We fix it by convention, choosing the origin so that the wrap of the phase falls
outside the projected image, where a smooth model could not represent it.

\subsection{Spatio-Temporal Self-Calibration}
\label{sec:selfcalib}

The system is calibrated from the same recording that transmits the temporal
code, without a reference object of known geometry.
We assume, however, that the intrinsic parameters of the cameras and of the
projector can be initialized from the design specification or from a calibration
carried out beforehand; these are initial values only, refined by the bundle
adjustment described below.
We assume in addition that the time map of \cref{sec:timemap} is already
calibrated, because the offset between the clocks is measured against it.

\heading{Projector-camera correspondences without calibration.}
The correspondence of \cref{sec:tempcode} is obtained from the identifier that
each dot transmits.
The decoding uses only the temporal structure of the events and no geometric
model of the camera or the projector, so the correspondence is available before
the system is calibrated, and for every feature the camera observes.
This property is what enables the self-calibration.

\heading{Geometric calibration.}
The correspondences determine the extrinsic parameters between any two views
in closed form, and the planar and the non-planar case are distinguished
automatically from the conditioning of the estimate~\cite{hartley2004multiple}.
Once an initial calibration has been obtained for every pair of views, the pairs
are merged by motion averaging~\cite{jiang2013global}, and the reprojection
error is then minimized non-linearly by bundle adjustment up to a global scale
factor.
The adjustment refines the intrinsic parameters as well, including the lens
distortion of the projector, and for that it needs correspondences at more than
one depth.
A scene that spans a range of depths supplies them; one dominated by a single
plane does not, and that is exactly the case the closed-form step detects and
for which it returns the plane normal.
We then split the correspondences by their height above that plane and draw
equal numbers from the two, because the plane otherwise outnumbers the rest
several times over and an adjustment that follows it leaves the plane correct
while displacing whatever stands in front of it by tens of millimeters.
If a metric reconstruction is required, it can be fixed by observing a reference
object of known size.

\heading{Time synchronization.}
The decoding of \cref{sec:tempcode} provides, for every dot, the projector pixel
and the phase of the dot inside the refresh period.
The time map of \cref{sec:timemap} gives the time at which that projector pixel
is illuminated.
The difference between the two is the offset between the clock of the camera and
the clock of the projector, and it is obtained from the same decoding that
provides the correspondences.
No additional recording and no hardware trigger are required
(\cref{fig:timeline}a).

The offset is determined relative to the convention of \cref{sec:timemap} that
fixes the constant term of the time map, not relative to the instant at which the
projector actually emits light, and measuring the latter would need an
independent optical reference.
The offset between two cameras is the difference of their offsets and does not
depend on that convention at all.
This is what the measurement of \cref{sec:capture} uses.

\subsection{3D Capture Algorithm}
\label{sec:capture}

3D measurement assumes that every projector and every camera is synchronized in
time.
It is then carried out by searching, under the epipolar constraint, for
correspondences between the projector pixels and the camera pixels that were lit
at the same instant~\cite{Muglikar213DV}, or by a lookup that replaces that
search~\cite{morgenstern2023x}.
To produce events over the whole scene the projector displays uniform white, so
it also serves as the illumination.
Uniform white carries no identifier, however, and by itself it cannot establish
the synchronization.
Each capture session therefore begins with a preamble, in which the projector
transmits the temporal code of \cref{sec:tempcode}.

\heading{Calibration preamble.}
The projector transmits the temporal code and we decode it in every camera.
The decoded dots yield two quantities.
The first is a set of projector-to-camera correspondences, which we use for the
self-calibration of \cref{sec:selfcalib}.
The second is the phase of each dot inside the refresh period, and comparing that
phase with the time map of \cref{sec:timemap} at the projector pixel of the dot
recovers the offset between the camera clock and the projector, because the two
refer to the same instant.
The refresh period of a free-running projector drifts, as \cref{sec:eval-sync}
measures, so this offset is re-estimated on the segment being measured.

\heading{Measurement phase.}
Ideally the event camera would resolve the instant at which the projector's spot
sweeps a point of the scene, a single camera pixel would fire, and the point
could be triangulated at once.
The temporal resolution of a real event camera is one microsecond, and within one
microsecond the projector advances by about 74
pixels.\footnote{An example, obtained from the projector described in
\cref{sec:setup}, which scans its $1280 \times 720$ pixels in about 12.4\,ms of
the 16.7\,ms frame. The value depends on the projector, but it is of this order
for any laser-scanning projector that refreshes at a video rate.}
A comparable number of pixels fires within the same microsecond on the camera
side, so one timestamp does not identify one pixel: what it determines is a
correspondence between a set of projector pixels and a set of camera pixels,
that is, between two line segments.

The microsecond is moreover only the smallest unit of the timestamp that is
recorded with each event, and events are not emitted exactly at the instant it
records.
The sensor arbitrates the readout by row, and the pixels of the row that is
acknowledged emit together and are given the same
timestamp~\cite{finateu2020back,lopes2026characterization}.
\Cref{sec:eval-timemap} reports how many event pixels actually share one
timestamp, and shows that this number varies both across the image and, for one
pixel observing a static scene, from one frame to the next.

The timestamp therefore provides a segment-to-segment correspondence, within
which we search for point correspondences under the epipolar constraint.
To account for its ambiguity we aggregate a few adjacent timestamps, over a
window well inside one line period.
The segment that the projector contributes is the preimage $\phi^{-1}(t)$ of the
time map of \cref{eq:timemap}, that is, the set of projector pixels that are lit
at the instant $t$.
\Cref{sec:eval-timemap} evaluates the resulting reconstruction.

\heading{Degenerate case.}
When the epipolar line and the scan line are close to parallel they do not
intersect at a well-defined point, which constrains the scan direction of the
projector and the placement of the cameras.

\section{Evaluations}

This section evaluates the parts of the method in the order in which they are
used.
\Cref{sec:eval-sync} measures how well the preamble of \cref{sec:capture}
synchronizes a camera to the projector, \cref{sec:eval-timemap} measures how far
the real scan departs from an ideal raster and what the departure costs in depth,
and \cref{sec:eval-selfcalib} evaluates the self-calibration.

The system under evaluation is one camera and one projector, which is the
minimal configuration of a structured light system and the configuration for
which the time map has to be calibrated, because the projector acts as a view.
Two cameras give a reconstruction that does not use the time map at all, and
that reconstruction is the reference the measurements are compared against; the
second camera of \cref{sec:setup} enters the measurement itself only through
the time origin, which is one scalar.

\subsection{Capture Setup}
\label{sec:setup}

The system consists of one laser-scanning projector and one event camera, with a
second event camera added for the evaluation only (\cref{fig:overview}).
The projector is an Ultimems HD309D1-C of $1280 \times 720$ pixels and a nominal
refresh rate of 60\,Hz, whose specification states a horizontal field of view of
$39^\circ$ and a brightness of 34\,lm at a projection distance of 400\,mm.
Both cameras are CenturyArks SilkyEvCam HD units with a $1280 \times 720$ Sony
IMX636 sensor, equipped with an 8\,mm lens.
The cameras were configured so that only positive events are produced.
Their intrinsic parameters were calibrated
beforehand~\cite{zhang2000flexible} and are refined by the self-calibration of
\cref{sec:selfcalib}, which runs in the preamble of \cref{sec:capture}.
The two cameras are synchronized in hardware.
The reconstruction is determined only up to the global scale factor that
\cref{sec:selfcalib} leaves free, so lengths are reported in units of $d$, the
distance from the measuring camera to the plane that the pattern is projected
onto, until \cref{sec:eval-timemap} fixes $d$ from an object of known size.

\begin{figure}[t]
  \centering
  \includegraphics[width=\linewidth]{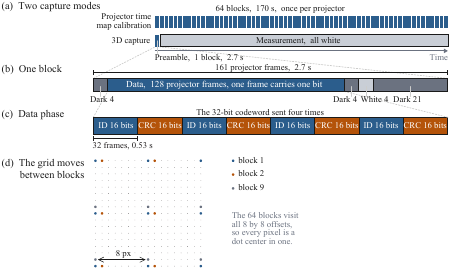}
  \caption{The pattern in time, at the four scales it has, and the grid it
    carries.
    (a) The two ways of capturing. The time map is calibrated once per
    projector, from a recording of 64 blocks; a 3D capture sends one block as
    the preamble of \cref{sec:capture} and then projects white while it
    measures.
    (b) One block, as the event rate of a recording measures it. The data phase
    is what the decoding of \cref{sec:tempcode} reads, the dark phases delimit
    it, and the all-white phase is the one \cref{sec:eval-timemap} uses to
    measure the scan.
    (c) The data phase carries the 32-bit codeword four times, one bit per
    frame, which is what lets a window of the length of the codeword find it
    wherever it starts.
    (d) The dots of one block sit on a grid of pitch 8, drawn here over the
    projector pixels, and the grid moves by one pixel from block to block, so
    the 64 blocks place a dot on every projector pixel in turn.
  }
  \label{fig:timeline}
\end{figure}

\heading{Projected pattern.}
\Cref{fig:timeline} collects the pattern in time and the grid it carries.
The features of \cref{sec:tempcode} are square dots of $3 \times 3$ projector
pixels arranged on a grid with a pitch of 8 pixels, so that one block contains
$1280/8 \times 720/8 = 160 \times 90 = 14{,}400$ dots, and the 64 blocks shift
the grid through every offset of that pitch.
A smaller dot locates the projector pixel it stands for more precisely, and a
larger one is detected more reliably, because an event camera fires on a noisy
threshold and a dot that covers more camera pixels is missed less often.
We chose $3 \times 3$ experimentally between the two.
Their identifiers are 16-bit words concatenated with a 16-bit CRC-16/CCITT-FALSE,
so the codeword is 32 bits long, takes 0.53\,s to transmit, and is repeated four
times.
Two repetitions are enough for the two-vote rule of \cref{sec:tempcode}; we send
four because a dot whose image is weak can fail to decode in one of them.
This combination detects all patterns of up to three random bit errors, all
burst errors of up to 16 bits, and every pattern with an odd number of bit
errors.
A block therefore carries 128 sampled bits per dot, the codeword four times
over, and the sliding window of \cref{sec:tempcode} has $128 - 32 + 1 = 97$
positions in them.
The CRC accepts a random one with probability $2^{-16}$, so a single window would admit
a few hundred wrong identifiers per recording and the two-vote rule makes that
number negligible.

\heading{Recordings.}
Eight recordings are used, listed in \cref{tab:recordings}, all of them carrying
the pattern just described.
Three of them, $R_1$ to $R_3$, were made on one day without moving the rig, and
they carry the time map of \cref{sec:eval-timemap}, the residual of
\cref{sec:eval-sync} and the self-calibration of \cref{sec:eval-selfcalib}.
$R_4$ was made on the day of $S_3$ and carries the check against the hardware
lock in \cref{sec:eval-sync}; the time map of $S_3$ is fitted on it as well.
The other four, $S_1$ to $S_4$, are the capture sessions of
\cref{fig:raster-depth}, one scene each, and each is calibrated from its own
identifier blocks rather than from $R_1$.
In $S_1$ to $S_3$ the identifier blocks and the all-white phase are two
recordings of the same scene made one after the other, which is how a long
recording of a static scene is convenient to make.
$S_4$ is one recording instead, a preamble followed by the measurement, which is
the form of capture the method is designed for.
One block is enough for that preamble, as \cref{sec:eval-sync,sec:eval-selfcalib}
show; four were sent here for margin.

\begin{table}[tb]
  \caption{The recordings. The two middle columns give the number of
    identifier blocks and the length of the recording that carries them, and
    the length of the all-white recording. A block is one shift of the grid,
    and the 64 shifts
    of a long recording place a dot on every projector pixel in turn. $S_1$ and
    $S_2$ were recorded on the same day without moving the rig, so the blocks
    of $S_1$ calibrate both. $S_4$ holds its blocks and its measurement in one
    recording of 25\,s, of which the first 8\,s are the preamble.
  }
  \label{tab:recordings}
  \centering
  \setlength{\tabcolsep}{3pt}
  \begin{tabular}{@{}llccl@{}}
    \toprule
    & Scene & Blocks & White & Used in\\
    \midrule
    $R_1$ & flat wall & 64, 170\,s & --- & \cref{sec:eval-sync,sec:eval-timemap,sec:eval-selfcalib}\\
    $R_2$ & box on the floor & 64, 170\,s & --- & \cref{sec:eval-selfcalib}\\
    $R_3$ & a second box & 64, 170\,s & --- & \cref{sec:eval-selfcalib}\\
    $R_4$ & flat wall & 64, 168\,s & --- & \cref{sec:eval-sync}\\
    \midrule
    $S_1$ & flat wall & 64, 170\,s & 10\,s & \cref{fig:raster-depth}\\
    $S_2$ & sphere, 150\,mm & $S_1$'s & 10\,s & \cref{fig:raster-depth}\\
    $S_3$ & robot & 64, 170\,s & 12\,s & \cref{fig:raster-depth}\\
    $S_4$ & face and hand & 4, 8\,s & 17\,s & \cref{fig:raster-depth}\\
  \bottomrule
  \end{tabular}
\end{table}

\subsection{Temporal Synchronization}
\label{sec:eval-sync}

The preamble of \cref{sec:capture} estimates the offset between a camera and the
projector as $\tau - \phi(p)$, so its error is the error of the timestamps
together with that of the time map, and this section separates the two.
The recording used here is $R_1$ of \cref{tab:recordings}, and the check
against the hardware lock below uses $R_4$.

Every decoded dot yields the phase at which a camera saw it, which is the offset
plus $\phi(p)$ if the time map is correct.
We remove one constant per block, the gauge freedom of \cref{sec:timemap},
and report what is left.
With the time map calibrated on this recording the residual is 6.83\,$\mu$s
standard deviation and 16\,$\mu$s at the 95th percentile, against a standard
error of 0.5\,$\mu$s for the phase of a single dot, so it is systematic rather
than measurement noise.

Its origin follows from comparing the two cameras on the 606,710 dots that both
decoded: the two residuals correlate at $+0.969$, the part they share has a
standard deviation of 6.71\,$\mu$s and the part private to one camera
1.67\,$\mu$s.
The shared part is a property of the projector and not of a camera: it is the
error of the cubic model of \cref{sec:timemap}, which \cref{sec:eval-timemap}
measures and propagates to depth.
The private part is the synchronization error itself, and averaging it over the
dots of one preamble reduces it to well below 0.1\,$\mu$s, so the length of the
preamble is set by the decoding of \cref{sec:tempcode}.

The two cameras are locked in hardware and the estimate never sees that lock, so
it can be checked against it.
Over the 64 blocks and 168\,s of $R_4$ the preamble reports an offset of
$-2.94\,\mu$s between them that repeats to 0.079\,$\mu$s, and a residual rate
difference of $-6.7 \times 10^{-10}$, which is the slope of that offset against
time and therefore the relative error between the two clock rates, where
free-running clocks would differ by parts per million.
The projector is free running and has no such reference, which is why its
offset is evaluated above through the residual instead.

One constant of the offset is not identifiable, because the constant term of the
time map absorbs it.
The reconstruction is unaffected, because it compares a timestamp with the time
map only after subtracting the offset, so a constant added to the one and taken
from the other cancels, in the convention of \cref{sec:timemap}.

We verified experimentally that one block recovers the camera-to-camera offset
to 0.045\,$\mu$s and 32 blocks to 0.031\,$\mu$s, so the temporal preamble
saturates at one block, as the geometric one does in \cref{sec:eval-selfcalib}.

The offset to the projector is a different matter, because the refresh period
is not constant: over the 168\,s of $R_4$ it falls from 16,693 to
16,690\,$\mu$s as the device warms up.
A dot's phase is folded over the 161 frames of its block, so a 1\,$\mu$s error
in the period displaces it by some 80\,$\mu$s and the offset a block reports is
tied to that block's own period estimate.
The time origin is therefore re-estimated on the segment being measured.

\subsection{Projector Time Map Calibration}
\label{sec:eval-timemap}

This section first measures what a timestamp of the sensor identifies, because
every measurement that follows is made on the events that share one timestamp.
It then establishes the three properties of the scan stated in
\cref{sec:timemap} and measures the error caused by the idealized raster.

\heading{Timestamps of the sensor.}
The projector advances about 74 pixels per microsecond, so if the sensor gave an
independent timestamp to every pixel the number of events carrying one timestamp
would be close to that value and would vary little.
Over the active part of the scan the mean is 59 events per timestamp, which is of
the expected order, but the standard deviation is 29 and only 37\,\% of the
timestamps carry a number within 25\,\% of the mean.
A resonant mirror does not sweep at a constant rate along a line, but
regressing this number on the phase within one scan line accounts for only
18\,\% of the variance.

A second measurement removes the projector from the comparison.
The scene is static and the illumination repeats every frame, so a pixel should
fire at the same instant in every frame and share its timestamp with the same
number of others.
Following 600 pixels over 20 frames, every pixel fires in almost every frame and
the instant repeats to 8.1\,$\mu$s, which is about half a scan line, but the
number of events sharing its timestamp has a mean of 73.6 and a standard
deviation of 32.1, ranges from 32 to 115 for a single pixel, and the phase within
the scan line now accounts for 0.1\,\% of the variance.
The sensor produces this variation, and the mechanism is the
readout~\cite{finateu2020back,lopes2026characterization}.
A timestamp therefore identifies a segment of the scan line rather than a point
on it, and the measurement of \cref{sec:capture} is based on this fact.

\begin{figure}[tb]
  \centering
  \makebox[\linewidth]{%
    \includegraphics[height=2.2cm]{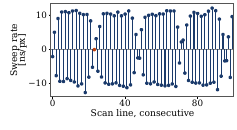}\hfill
    \includegraphics[height=2.2cm]{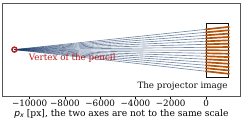}}
  \caption{The structure of the projector scan, measured under a full-white
    projection.
    Left: the rate at which the event time advances along each of 100
    consecutive scan lines. It changes sign from one line to the next, which is
    the alternation. The one line drawn in orange is a window that straddles a
    line boundary, where the rate is not significant.
    Right: the iso-timestamp contours of the calibrated time map, in orange
    inside the projector image, and their extensions, which meet at the vertex
    of a pencil far outside it. A parallel raster would put these contours
    parallel and leave no vertex. The test is made in the coordinates of the
    projector, where an oblique view of the scene cannot produce the
    convergence.
  }
  \label{fig:scan}
\end{figure}

\heading{Alternation of the scan direction.}
Under a full-white projection we cut the event stream into windows one line
period long, fitted a straight line to the events of each in the image plane,
and regressed event time against position along it.
Over one white phase of 63.5\,ms, 2,991 windows carried enough events for a fit
and 2,827 gave a slope significant at $3\sigma$, 1,412 scanning left to right
and 1,415 right to left, and 98.9\,\% of the 2,703 adjacent pairs were scanned
in opposite directions (\cref{fig:scan}, left).
This is the expected behavior of a resonant mirror and it excludes the
unidirectional raster.

The alternation does not have to be modeled, for the reasons given in
\cref{sec:timemap}.

\heading{Fan geometry of the scan lines.}
The same line fits yield the tilt of each scan line, which runs from
$-2.45^\circ$ to $+2.45^\circ$, a change of $4.90^\circ$, so the lines are not
parallel.
The point that minimizes the sum of the squared distances to all of them is the
vertex of the pencil, and it lies 6.7 frame widths outside the image, at
$0.46$ of the height of the frame.
A pencil of lines in the camera is not by itself evidence of a fan in the
projector, because a parallel raster observed obliquely also converges, at the
vanishing point of the direction of the lines.
We therefore repeat the test in projector coordinates alone, on the
iso-timestamp contours of the fitted time map, whose tilt changes monotonically
from $-1.40^\circ$ at the top of the frame, through $0^\circ$ near
$p_y \approx 350$, to $+1.50^\circ$ at the bottom, where a parallel raster would
give $0.00^\circ$ everywhere (\cref{fig:scan}, right).
Extended, those contours meet at $(-10.9 \times 10^{3},\, 350)$ in projector
pixels, and mapping the camera-side vertex through the calibration plane places
it at $(-10.1 \times 10^{3},\, 365)$.
The two estimates use disjoint information, the first the raw events and the
camera calibration and the second only the fitted time map, and they agree on
the side, the order of magnitude and the height, which is also where the tilt of
the contours vanishes.
The scan is therefore a fan.

\heading{Origin of the fan.}
The fast axis of these projectors is a mirror at resonance and the slow axis a
sawtooth at the refresh rate driven off
resonance~\cite{chen2026dualaxis,ultimems}, so the slow axis keeps moving while
a line is written and the line is not horizontal.
Straight rows at a constant spacing are reported for one design only, which
steps the slow axis under position feedback and whose authors describe that as
what separates them from scanners in which ``projection is performed at an
angle (rather than a straight line)''~\cite{hornstein2010enhanced}.
A constant tilt would still leave the lines parallel.
What can make them converge is that the two mirror rotations do not commute, so
the map from the pair of mirror angles to a flat screen is not separable.
A single biaxial mirror with an obliquely incident beam projects a surface whose
intersection with the screen is neither rectilinear nor symmetric about the
frame, and that distortion, like the sinusoidal rate along a line, is
compensated by choosing when each pixel is
emitted~\cite{hornstein2012maradin}.
Since $\phi$ is defined on the pixels as emitted, what the compensation leaves
behind is what $\phi$ records.
We did not open the projector, and give this as the mechanism that accounts for
the measurement rather than as a measured cause.

\heading{The same fan on three units.}
\Cref{tab:fan-units} repeats the measurement on the two further units we hold,
mounted together with the first in a later rig and calibrated there from
recordings of the same temporal code.
All three place the vertex on the same side of the image, 7.9 to 8.5 frame
widths out and between 0.38 and 0.53 of the frame height, and depart from the
best affine ramp by 117 to 127\,$\mu$s rms.
The shape of the fan repeats across units while its parameters do not, which is
the case for calibrating $\phi$ rather than assuming any fixed form for it, the
idealized raster included.
The last row is unit $A$ calibrated from $R_1$ instead of in the later rig, so
it and the first row are one device measured in two rigs.
They agree on the vertex and differ by 0.4$^\circ$ in the tilt at the edges,
which is the precision the later rig allows: its fits reach 23 to
25\,$\mu$s rms rather than the 5\,$\mu$s of \cref{tab:timemap-models},
because it carries camera intrinsics measured on another body.
That is still well below the 500\,$\mu$s the fan spans.

\begin{table}[tb]
  \caption{The fan on the three units of the projector we hold. $A$ is the one
    used throughout the paper. Each iso-timestamp contour of the fitted $\phi$
    is traced across the projector image and fitted by a straight line, as in
    \cref{fig:scan}; the vertex is where those lines meet, in frame widths
    outside the image. The last column is how far the fitted $\phi$ departs
    from the best affine ramp over the whole projector image, which is near but
    not equal to the fit residual of \cref{tab:timemap-models}. The last row
    is $A$ calibrated from the dense recording $R_1$ of \cref{tab:recordings}
    instead of in the later rig, which is the same device measured twice.
  }
  \label{tab:fan-units}
  \centering
  \setlength{\tabcolsep}{4pt}
  \begin{tabular}{@{}lcrr@{}}
    \toprule
    Unit & Tilt, edge to edge & Vertex & Affine ramp\\
         & [deg] & [widths] & [$\mu$s rms]\\
    \midrule
    $A$ & $-1.00$ to $+1.93$ & 8.5 & 117.1\\
    $B$ & $-1.52$ to $+1.63$ & 7.9 & 127.4\\
    $C$ & $-1.66$ to $+1.42$ & 8.1 & 122.6\\
    \midrule
    $A$, from \cref{sec:setup} & $-1.41$ to $+1.49$ & 8.5 & 118.5\\
  \bottomrule
  \end{tabular}
\end{table}

\heading{Error caused by the idealized raster.}
\Cref{tab:timemap-models} compares models of $\phi$ on the same 684,079 decoded
dots, each with one constant per block as \cref{sec:timemap} describes.
The recording is $R_1$ of \cref{tab:recordings}, whose 64 blocks cover the
projector image, and not the one-block preamble of \cref{sec:capture}.

\begin{table}[tb]
  \caption{Time map models, fitted to the same 684,079 decoded dots.
    The idealized raster is the affine ramp $a + b\,p_x + c\,p_y$.
    The fan model is a polynomial in the angle at the pencil vertex: degree 1
    carries one coefficient and degree 3 three, and each carries a constant and
    the two coordinates of the vertex as well, which is where the 4 and the 6
    come from. The free constant per block is common to every model and is not
    counted.
  }
  \label{tab:timemap-models}
  \centering
  \setlength{\tabcolsep}{4pt}
  \begin{tabular}{@{}lrr@{}}
    \toprule
    Model of $\phi$ & Parameters & Residual rms [$\mu$s]\\
    \midrule
    Affine ramp (idealized raster) & 3 & 119.34\\
    Bivariate quadratic & 6 & 27.48\\
    \textbf{Bivariate cubic (ours)} & \textbf{10} & \textbf{5.07}\\
    \midrule
    Fan: degree 1 in the angle at the vertex & 4 & 31.79\\
    Fan: degree 3 in the angle & 6 & 31.52\\
  \bottomrule
  \end{tabular}
\end{table}

Two results follow from this table.
First, the error of the idealized raster is 119\,$\mu$s rms, and most of it is
caused by the fan: parameterizing $\phi$ by the angle at the vertex of the
pencil brings the residual down to 32\,$\mu$s with four parameters.
The fan is therefore the dominant deviation.
Second, the fan does not explain the whole deviation: the angle model saturates
at 32\,$\mu$s when more terms are added, while the cubic polynomial reaches
5.07\,$\mu$s.
The angle model cannot represent the timing inside one line, because a resonant
mirror scans sinusoidally and the time is not linear in the position along it,
nor the remaining non-linearity of the slow axis.
For these reasons we use the bivariate cubic polynomial of \cref{eq:cubic}
rather than the fan model.

\heading{Against a map measured on a plane.}
\Cref{sec:related} describes how X-maps calibrates $\phi$, by projecting white
onto a plane and carrying the camera's timestamps into the projector image by a
homography fitted to four corners.
We repeated it on an all-white recording of the wall from the session of $R_1$,
rectifying the camera first because its distortion moves a corner by 20 pixels,
and scored it on the 635,848 dots of $R_1$ it covers, one constant per block.
It leaves 43.5\,$\mu$s rms, against 116.4\,$\mu$s for the affine ramp and
5.05\,$\mu$s for our cubic scored the same way, as a fixed map and not a fit.
Measuring the map is therefore worth a factor of 2.7 over assuming the raster,
and calibrating it as we do a further 8.6.
Two things limit it.
Fitting the homography to all the decoded correspondences instead of to four
corners, which a method that projects only white does not have, brings the
residual to 31.0\,$\mu$s, and even then it is 2.8 pixels from them, which is
about 50\,$\mu$s at the 17.5\,$\mu$s per row at which $\phi$ advances.
Reading the camera-side map at the dot's own camera pixel, with no warp at all,
still leaves 20.7\,$\mu$s, because one timestamp is shared by many pixels and
by a number of them that varies (\cref{sec:eval-timemap}), so a map measured on
the camera grid carries the sensor into a quantity that describes the
projector.
Our calibration avoids both, because the temporal code names the projector
pixel of each dot outright.

\begin{figure}[t]
  \centering
  \includegraphics[width=\linewidth]{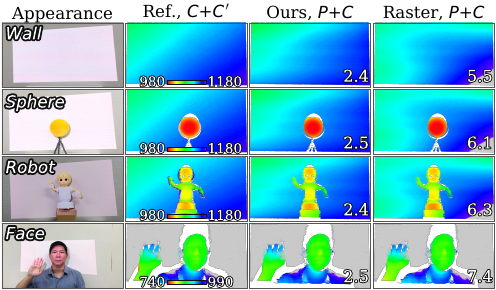}
  \caption{Quantitative evaluation of 3D shape reconstruction.
    Each row is one recording of one object.
    The first column is a photograph of the object, and the other three are
    depth maps rendered in the pixels of the measuring camera $C$: the
    reference reconstruction, from the two cameras, which does not use the time
    map at all; the measurement under test, from the projector and $C$ with the
    calibrated time map; and the same measurement with the time map replaced by
    the best-fitting ramp, the idealized raster.
    The number in the corner is the median absolute difference from the
    reference, and the bar of the reference panel gives the color scale; both
    are in millimeters.
    The three depth maps of a row share that scale.
    Gray is depth outside the window, white the absence of a measurement.
    We can observe that the idealized raster bends the flat wall into a
    saddle.
  }
  \label{fig:raster-depth}
\end{figure}

We finally measure the error in depth, on correspondences taken from an
all-white projection, because that is what gives a depth at every pixel.
We reconstruct each of the four capture sessions $S_1$ to $S_4$ twice from the
projector and the measuring camera, once with the calibrated time map and once with the ramp that
fits it best, and compare both against the reference reconstruction from the two
cameras (\cref{fig:raster-depth}).
The correspondences of the other view are pooled over $\pm 3\,\mu$s as described
in \cref{sec:capture}, which covers between 64 and 83\,\% of the pixels of the
measuring camera, the lower value corresponding to the recording of a person,
whose hair reflects little light.
Expressed in the millimeters that the sphere of known diameter fixes below, the
calibrated map agrees with the reference to between 2.4 and 2.5\,mm in median
absolute depth, and the idealized raster to between 5.5 and 7.4\,mm.
The idealized raster is therefore between two and three times worse, and the
error it makes is not noise but the structure of \cref{fig:overview}, right,
carried into depth.
That structure is weak at the center of the frame and grows outwards: within a
quarter of the way from the center to a corner the raster is 2.5\,mm from the
reference against 2.0\,mm for the calibrated map, whereas in the outermost
quarter it is 12.9\,mm against 3.5\,mm.
A measurement confined to the middle of the field would therefore see little
difference; one that fills the frame with the object sees it at the edges.

\heading{Flatness of the wall.}
The comparison above is against a reconstruction that shares its geometric
calibration with the measurement under test, so an error common to the two
would cancel in their difference.
The wall settles that, because its flatness is a property of the scene and not
of any reconstruction.
We fit a plane to each reconstruction of the wall and report what is left of it
(\cref{tab:wall}).
The reference is 1.51\,mm rms from its own plane, which bounds how far the wall
itself departs from flat.
The calibrated time map gives 3.07\,mm and the idealized raster 8.88\,mm, so on
a criterion that no reconstruction defines the idealized raster is 2.9 times
worse, and what it adds is the saddle of \cref{fig:raster-depth}.
The growth toward the edges appears here as well.
Within a quarter of the way from the center of the frame to a corner the two
are 1.77 and 2.95\,mm, and in the outermost quarter 4.98 and 20.93\,mm.
The sphere below is held at one position in the frame, whereas the wall fills
it, so the two cover the middle of the field and its edges between them.

\begin{table}[tb]
  \caption{Residual of a plane fitted to the reconstructed wall, which needs no
    reference reconstruction. The wall spans 851 by 496\,mm and the fit uses
    7,402,626 points; the 505 points that lie more than 50\,mm from the plane
    of the reference are discarded as outliers, once, for all three columns.
    The regions are measured from the center of the frame toward a corner, and
    the percentile is of the absolute residual.
  }
  \label{tab:wall}
  \centering
  \setlength{\tabcolsep}{4pt}
  \begin{tabular}{@{}lrrr@{}}
    \toprule
    Residual [mm] & Reference & Calibrated & Idealized\\
                  & $C$+$C^\prime$ & map, $P$+$C$ & raster, $P$+$C$\\
    \midrule
    Whole frame, rms   & 1.51 & \textbf{3.07} & 8.88\\
    Whole frame, 90th percentile & 1.73 & \textbf{5.06} & 13.65\\
    Inner quarter, rms & 1.22 & \textbf{1.77} & 2.95\\
    Outer quarter, rms & 2.28 & \textbf{4.98} & 20.93\\
  \bottomrule
  \end{tabular}
\end{table}

$S_4$ is the form of capture the method is designed for, a preamble followed by
the measurement in one recording, and it is calibrated from that preamble
alone; the person raises a hand, which the measurement resolves at the same
time as the face.

One of the objects is a sphere of known diameter, which is an external
reference and not another reconstruction.
Fitting a sphere to the reference reconstruction alone fixes its center and its
radius, and the known diameter of 150\,mm turns that radius into the scale that
\cref{sec:selfcalib} leaves free, here $d = 1127\,$mm.
Held at that position and that size, the sphere is 1.10\,mm rms from the
reference that defined it, $1.39\,$mm rms from the measurement made with the
projector and one camera, and $1.72\,$mm rms from the same measurement under
the idealized raster.

\subsection{Self-Calibration}
\label{sec:eval-selfcalib}

The temporal part of the self-calibration of \cref{sec:selfcalib} is measured in
\cref{sec:eval-sync}, so this section covers the geometric part.
No external reference for the extrinsic parameters is available, so we report how
the estimate repeats over independent recordings and how much of the preamble it
needs.
The three recordings $R_1$ to $R_3$ of \cref{tab:recordings} are used, and the
correspondences are the decoded dots, about 600,000 in each.

\heading{Repeatability.}
$R_1$ to $R_3$ were made without moving the rig, so their solutions should
agree.
They agree to $0.11^\circ$ in relative rotation, to $0.08^\circ$ in the normal of
the dominant plane, and to 1.0\,\% in $|t|/d$, the ratio of the baseline to the
distance to that plane, whose value is 0.281.
The two recordings that contain a box are solved by the same route, because the
wall behind the box is still the dominant plane, and the share of correspondences
the homography explains falls from 98.7\,\% to 85.9 and 74.9\,\%, which is the
box.
Reconstructing the wall from the two cameras with these parameters leaves a plane
whose thickness is $1.25 \times 10^{-3}\,d$.

\heading{Length of the preamble.}
\Cref{tab:preamble} compares solutions computed from prefixes of one recording
against the solution computed from the whole of it.
One block, which is 2.7\,s, already reaches the accuracy that 32 blocks reach, so
the estimate is limited by something other than the number of correspondences.
Subsampling inside a single block gives the same answer: 50 of its 10,882
correspondences still give $0.05^\circ$ in rotation, $0.10^\circ$ in baseline
direction and 0.23\,\% in $|t|/d$.
The same holds on the recordings a solution was not estimated from: one block
reaches 0.239\,px and all 64 reach 0.246\,px.
The length of the preamble is therefore set by the time the codeword takes to
transmit and not by the accuracy of the calibration.
The codeword takes 0.53\,s and is repeated four times here; two repetitions
suffice for the two-vote rule of \cref{sec:tempcode}, which would bring the
preamble to 1.1\,s.

\begin{table}[tb]
  \caption{Self-calibration from a prefix of the preamble, compared against the
    solution from all 64 blocks of the same recording. Median over 32 random
    prefixes of each length; the 90th percentile of an entry is at most four
    times its median.
  }
  \label{tab:preamble}
  \centering
  \setlength{\tabcolsep}{3pt}
  \begin{tabular}{@{}lrrrrrr@{}}
    \toprule
    Blocks & 1 & 2 & 4 & 8 & 16 & 32\\
    \midrule
    Time [s]                    & 2.7 & 5.4 & 10.8 & 21.5
                                & 43.0 & 86.1\\
    Corresp.\ [$10^3$]          & 10.9 & 21.6 & 43.5 & 85.3 & 174 & 328\\
    Rotation [$^\circ$]         & 0.018 & 0.023 & 0.018 & 0.019
                                & 0.019 & 0.019\\
    Baseline [$^\circ$]         & 0.035 & 0.051 & 0.034 & 0.043
                                & 0.049 & 0.052\\
    $|t|/d$ [\%]                & 0.068 & 0.067 & 0.084 & 0.072
                                & 0.073 & 0.037\\
    \bottomrule
  \end{tabular}
\end{table}

\section{Conclusion}
We have calibrated the function that returns the instant at which a
laser-scanning projector illuminates each of its pixels, and we have measured
the error incurred when that function is assumed instead.
On our system the real scan departs from an idealized raster by up to
457\,$\mu$s, and a reconstruction that assumes the raster is two to three times
less accurate in depth than one that uses the calibrated map.
The error is a structured deformation of the scene rather than noise.
The sensor deviates from its own idealized model as well: one timestamp is
shared by many pixels, so it identifies a segment of the scan line and not a
point on it, and the measurement is designed accordingly.
The geometry and the clocks are recovered together from the projected code
itself, as a preamble that needs neither a reference object nor a second camera
and that takes a few seconds.

\bibliographystyle{ieicetr}
\bibliography{main}

\end{document}